%% file: main.tex
\documentclass[sigconf]{acmart}
\acmSubmissionID{1105}

\usepackage{booktabs} 

\usepackage[ruled]{algorithm2e} 

\SetAlFnt{\small}
\SetAlCapFnt{\small}
\SetAlCapNameFnt{\small}
\SetAlCapHSkip{0pt}

\input{preamble}

\begin{document}
\title[\moniker]{\moniker: Large-scale Consistent Street View Generation Using Autoregressive Video Diffusion}

\author{Boyang Deng}
\affiliation{
\institution{Stanford University}
\country{USA}
}
\email{bydeng@stanford.edu}
\author{Richard Tucker}
\affiliation{
\institution{Google Research}
\country{USA}
}
\email{richardt@google.com}
\author{Zhengqi Li}
\affiliation{
\institution{Google Research}
\country{USA}
}
\email{zhengqili@google.com}
\author{Leonidas Guibas}
\affiliation{
\institution{Stanford University}
\institution{Google Research}
\country{USA}
}
\email{guibas@cs.stanford.edu}
\author{Noah Snavely*}
\affiliation{
\institution{Google Research}
\country{USA}
}
\email{snavely@google.com}
\author{Gordon Wetzstein*}
\affiliation{
\institution{Stanford University}
\country{USA}
}
\email{gordon.wetzstein@stanford.edu}

\renewcommand\shortauthors{Deng, B. et al}
\begin{abstract}
We present a method for generating \emph{\artifacts}---long sequences of views through an on-the-fly synthesized city-scale scene.
Our generation is conditioned by language input (e.g., city name, weather), as well as an underlying map/layout hosting the desired trajectory.
Compared to recent models for video generation or 3D view synthesis, our method can scale to much longer-range camera trajectories, spanning several city blocks, while maintaining visual quality and consistency.
To achieve this goal, we build on recent work on video diffusion, used within an autoregressive framework that can easily scale to long sequences.
In particular, we introduce a new \emph{temporal imputation} method that prevents our autoregressive approach from drifting from the distribution of realistic city imagery. 
We train our \moniker system on a compelling source of data---posed imagery from Google Street View, along with contextual map data---which allows users to generate city views conditioned on any desired city layout, with controllable camera poses.
\blfootnote{* Equal Contributions}
\end{abstract}
\input{figs/teaser}

%
%

%
%


\maketitle

\input{secs/01_intro}
\input{secs/02_rel_work}
\input{secs/03_method}
\input{secs/04_results}
\input{secs/05_conclusion}

\bibliographystyle{ACM-Reference-Format}
\bibliography{references}

\input{figs/long_track}
\input{figs/autoreg_syn}
\input{figs/text_control}
\input{figs/loc_mixmatch}

\input{supplement}

\end{document}

%% file: preamble.tex
\usepackage{microtype}
\usepackage{enumitem}
\usepackage{bbding}
\usepackage{wrapfig}
\usepackage{multirow}
\usepackage{xspace}
\usepackage{colortbl}
\usepackage{hyperref}

\usepackage{color}
\definecolor{turquoise}{cmyk}{0.65,0,0.1,0.3}
\definecolor{purple}{rgb}{0.65,0,0.65}
\definecolor{dark_green}{rgb}{0, 0.5, 0}
\definecolor{orange}{rgb}{0.8, 0.6, 0.2}
\definecolor{red}{rgb}{0.8, 0.2, 0.2}
\definecolor{darkred}{rgb}{0.6, 0.1, 0.05}
\definecolor{blueish}{rgb}{0.0, 0.3, .6}
\definecolor{light_gray}{rgb}{0.7, 0.7, .7}
\definecolor{pink}{rgb}{1, 0, 1}
\definecolor{greyblue}{rgb}{0.25, 0.25, 1}
\definecolor{tbl_no1}{rgb}{0.42, 0.76, 0.21}
\definecolor{tbl_no2}{rgb}{0.68, 0.87, 0.51}

\newcommand\blfootnote[1]{%
  \begingroup
  \renewcommand\thefootnote{}\footnote{#1}%
  \addtocounter{footnote}{-1}%
  \endgroup
}

\newcommand{\Fig}[1]{Fig.~\ref{fig:#1}}

\newcommand{\Tab}[1]{Tab.~\ref{tbl:#1}}

\newcommand{\Eq}[1]{Eq.~\eqref{eq:#1}}

\newcommand{\Sec}[1]{Sec.~\ref{sec:#1}}

\newcommand{\gauss}{\mathcal{N}}

\SetKwComment{Comment}{$\triangleright$\ }{}

\makeatletter
\DeclareRobustCommand\onedot{\futurelet\@let@token\@onedot}
\def\@onedot{\ifx\@let@token.\else.\null\fi\xspace}
 
\def\ie{\emph{i.e}\onedot}

\makeatother

\newcommand{\moniker}{Streetscapes\xspace}
\newcommand{\artifact}{Streetscape\xspace}
\newcommand{\artifacts}{Streetscapes\xspace}

\copyrightyear{2024}
\acmYear{2024}
\setcopyright{acmlicensed}\acmConference[SIGGRAPH Conference Papers '24]{Special Interest Group on Computer Graphics and Interactive Techniques Conference Conference Papers '24}{July 27-August 1, 2024}{Denver, CO, USA}
\acmBooktitle{Special Interest Group on Computer Graphics and Interactive Techniques Conference Conference Papers '24 (SIGGRAPH Conference Papers '24), July 27-August 1, 2024, Denver, CO, USA}
\acmDOI{10.1145/3641519.3657513}
\acmISBN{979-8-4007-0525-0/24/07}

%% file: figs/teaser.tex
\begin{teaserfigure}
\centering
\includegraphics[width=\linewidth]{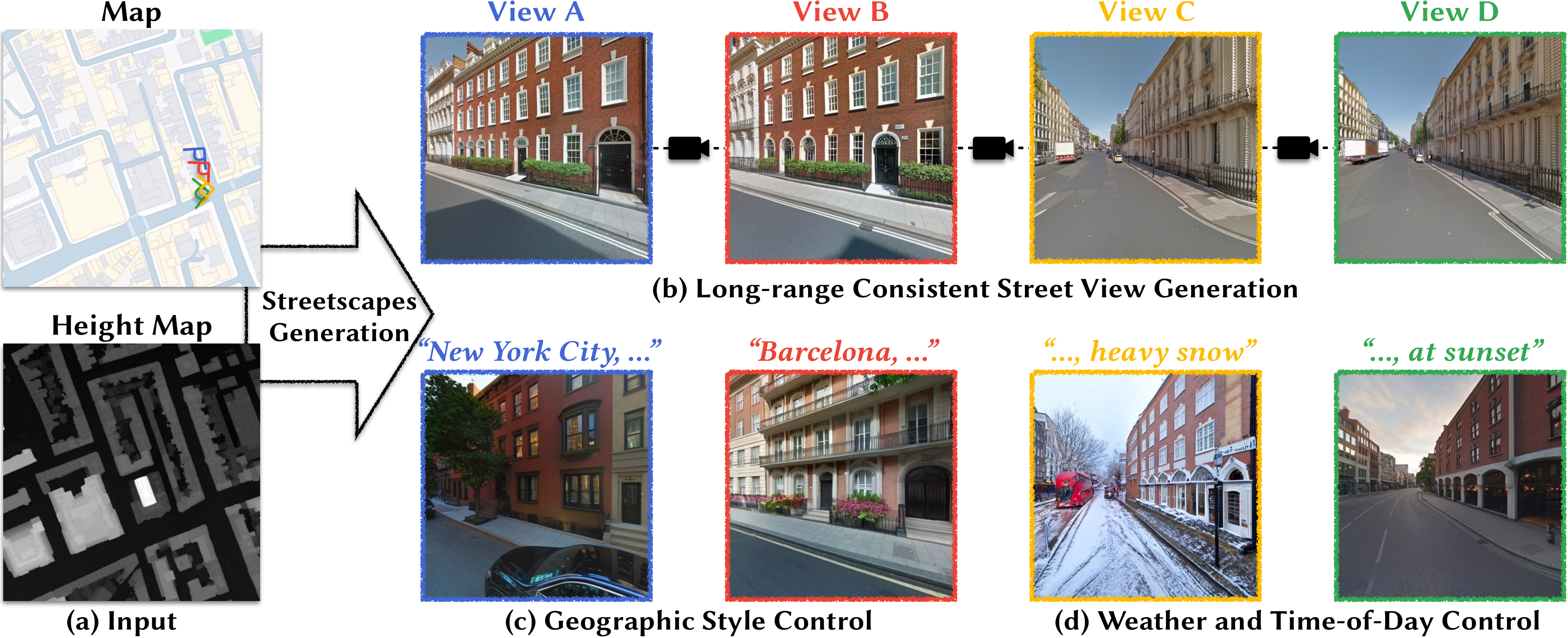}
\caption{
\textbf{Generating \artifacts}.
We design a system that generates \emph{\moniker}, i.e., consistent street views along long-range paths through large-scale synthesised urban scenes.
Given a)~a street map and its corresponding height map, our system can generate:
b)~high-quality street views spanning a long camera path with layout controlled by the map,
with the ability to generate c)~different geographic styles as well as 
d)~different weather conditions or times of day,
both controlled by text prompts.
We recommend the reader to visit our project page at \href{https://boyangdeng.com/streetscapes/}{\color{magenta}{boyangdeng.com/streetscapes}} to view our results as videos.
}
\label{fig:teaser}
\end{teaserfigure}

%% file: secs/01_intro.tex
\section{Introduction}
\label{sec:intro}

Methods for generating visual contents like images, videos, and 3D models have captured the world's imagination~\cite{po2023state}.
While the results are astounding, these methods often struggle when asked to produce large-scale outputs.
For instance, current text-to-video methods are limited to generating short videos of scenes with bounded extent. Text-to-3D methods can generate individual objects~\cite{pooledreamfusion}, but not whole cities.
In this work, we focus on the domain of urban scenes, and generate long-range, 3D-consistent views along paths through synthesized city streets, an output we call \emph{\artifacts}.

There are several challenges associated with generating outputs at the scale of \artifacts:
\begin{itemize}
  \item
  \textbf{What input to use?} Text is a popular form of conditioning. But at the scale of a city, text is a blunt tool that does not allow fine-grained control over the output. Instead, we condition our \artifacts on an overhead \emph{scene layout}, consisting of a street map and corresponding height map, inspired by recent layout-conditioned generative models~\cite{bahmani2023cc3d}. The unique combination of our input data allows us to render semantic labels, a height map, and a depth map from any camera pose, transforming the map--space information to screen space. Our method also takes in an optional text input describing the desired style, like weather and time of day.
  \item
  \textbf{What output to produce?} We want our outputs to be view consistent, as if they were rendered from a 3D model. But it is difficult to scale generative 3D methods to entire cities. Instead, we build on text-to-video models and produce a long sequence of images as output. However, current video generation methods cannot ensure long-range quality and 3D consistency. To address these issues, we ensure view consistency by conditioning each output image on the scene layout, projected to an image-space G-buffer~\cite{deering1988triangle}, and scale to long-range sequences via a new autoregressive video synthesis technique that minimizes drift away from the space of natural street scenes, while ensuring frame to frame continuity.
  \item
  \textbf{What training data to use?} Training our method requires large amounts of real street view sequences, along with corresponding scene layouts. Fortunately, such data has been captured at scale by mapping services like Google Street View~\cite{anguelov2010gsv}. 
  One of our key insights is to leverage such geographic datasets for generative tasks.
  However, there are also challenges associated with such data, like the fact that the scene layouts from mapping sites are relatively coarse-grained and not pixel-perfect when projected into ground views. 
  We show how to leverage mapping corpora like Google Street View for training, while achieving robustness to coarse-grained, noisy data.
\end{itemize}

In short, our work presents a new method for synthesizing long-range, consistent \artifacts by training on an overlooked source of data---large-scale collections of street view imagery---and conditioning on a new kind of input, namely scene layouts spanning multiple city blocks. To achieve high-quality \artifacts over long scales, we propose (i) a layout-conditioned generation approach, (ii) a motion module that enables consistent two-frame generation, and (iii) an autoregressive \emph{temporal imputation} technique that modifies the pre-trained two-frame motion module at inference time to enable consistent long-range video generation. With our results, we demonstrate that our system autoregressively generates \artifacts covering long-range camera trajectories with consistently high quality.
Our system also enables numerous creative scene generation applications, thanks to our flexible control over scene layout, camera poses, and scene conditions.

%% file: secs/02_rel_work.tex
\section{Related Work}
\label{sec:rel_work}

The generation of walkthroughs in large environments is a classic problem in computer graphics that motivated a lot of the early work in efficient visibility culling~\cite{teller,walkthroughs}. That line of works, however, assumes that a model of the environment is given, while our key focus is in the synthesis of views along a path through an environment, where the environment itself is created on the fly using modern generative neural methods, such as diffusion. A comprehensive review of the theory and practice of diffusion models for visual computing, 
as applied to 3D and video generation, can be found in the recent survey paper by Po et al.~\shortcite{po2023state}. We briefly cover the most relevant works here.

\paragraph{3D Object Generation}
Diffusion models have emerged as state-of-the-art generative models for 2D images~\cite{ho2020denoising, song2020score, rombach2022high}.
These models have also been extended to the 3D domain. For example, some strategies~\cite{pointe, shape, tridiff, 3dgen, auto3d} use direct 3D supervision. The quality and diversity of their results, however, is far from that achieved for 2D image generation.
DreamFusion~\cite{pooledreamfusion} and related works enable text-to-3D generation by combining a neural scene representation with the image priors of pre-trained 2D diffusion models ~\cite{magic3d,Chen_2023_ICCV,sjc,shi2023MVDream,3dgen}. These methods successfully generate 3D objects of moderately high quality. 
Image-conditioned 3D generation has also become a popular approach~\cite{watson2022novel,liu2023zero,genvs,nerfdiff,hong2023lrm}, with the latest methods in this category~\cite{hong2023lrm} resolving flicker and other view-inconsistent artifacts that plagued earlier approaches. 
Based on the idea of training 3D diffusion models directly on image data~\cite{holodiff,ssdnerf,shen2023gina}, state-of-the-art results for 3D generation have recently been achieved by techniques that solve the 3D generation problem using 2D diffusion models with 3D-aware denoisers~\cite{renderdiff,xu2023dmv3d}.

While different variants of these 3D diffusion approaches have become very successful, they are all limited to generating individual objects. This is in part because available training datasets, such as Objaverse~\cite{deitke2023objaversexl}, only contain individual objects; score distillation sampling (SDS)--based methods, such as DreamFusion, suffer from mode collapse that also effectively limits these methods to generating individual objects.

\paragraph{3D Scene Generation}
Large-scale and perpetual scene generation has been a topic of interest and much progress has been made~\cite{hao2021gancraft,liu2021infinite,li2022_infinite_nature_zero,chai2023persistentnature,lin2023infinicity,Yang2023}. Among these, Infinicity~\cite{lin2023infinicity} is closest to our work because they also aim for large-scale street view generation. Yet, these approaches build on generative adversarial networks (GANs) that are challenging to train and that have shown limited quality and diversity in practice. Moreover, the outputs of these GANs are not easily editable, for example by text prompts.

Emerging diffusion models have been used for 3D scene generation of rooms~\cite{hoellein2023text2room} and outdoor scenes~\cite{cai2022diffdreamer,SceneScape,shen2023make}. These methods iteratively apply depth estimation, depth-based image warping, and diffusion-based inpainting to generate a scene. Several very recent methods extend this approach to 3D Gaussian--based representations~\cite{ouyang2023text,chung2023luciddreamer}. Monocular depth estimation, however, is not always reliable, especially for street views, which makes it difficult to apply these methods to our setting. Moreover, a desired camera trajectory cannot easily be defined because the scene geometry is generated and not known a priori.

Another class of methods uses 3D bounding boxes to guide an object-centric diffusion model in generating a compositional scene \cite{Po2023Compositional3S,schult23controlroom3d}. While successful, these methods are not directly applicable to consistent 3D generation at a city scale and, like other SDS-based approaches, suffer from mode collapse and limited quality. CC3D also generates compositional room-sized scenes, albeit conditioned on a 2D image of a room's layout~\cite{bahmani2023cc3d}. CC3D employs a GAN architecture and inherits its limitations, namely limited editability and diversity. 

A set of recent, unpublished, and concurrent works discuss 3D street view generation \cite{swerdlow2023street,yang2023bevcontrol,gao2023magicdrive}. These methods take a bird's eye view---a layout map with each object in the scene---and ground truth 3D bounding boxes of all objects as input to generate street view images. Among these MagicDrive~\cite{gao2023magicdrive} achieves the highest-quality results and, similar to our approach, builds on a latent 2D diffusion model. 
Our work operates in a different setting: we do not have access to individual objects, such as cars, or their 3D bounding boxes.
Our maps only contain the layout of streets and buildings, along with their height. Moreover, their approach operates in a frame-by-frame manner and promotes consistency among frames using cross-frame attention; they discuss extensions to fine-tuning their model on a text-to-video model~\cite{wu2023tune}. 
In contrast, our autoregressive approach is not limited to the small numbers of frames that can be generated by existing text-to-video models.

Another very recent and concurrent approach to 3D scene generation is WonderJourney~\cite{yu2023wonderjourney}, where several keyframes are generated first and then interpolated using diffusion models. 

Unlike these approaches, the Streetscape setting is unique in 
offering ground truth proxy geometry of the buildings via the given map and corresponding height information. Simple as it may seem, this small amount of extra information enables unprecedented levels of control over the generation process, including controllable camera trajectories, 3D multi-view consistency, and coarse grounding of the generated scene layout for consistent long-range generation. Interestingly, our unique setting allows us to formulate the 3D generation process in a way that is more akin to video generation than to 3D generation.

\paragraph{Video Generation}
Video diffusion models ~\cite{singer2022make,ho2022video,blattmann2023align,blattmann2023stable,li2023generative,guo2023animatediff, villegas2022phenaki}
have recently emerged as data-driven approaches to training foundation models that can generate short video clips of impressive quality and consistency. ControlNets can be used to condition the video generation process on depth maps or human poses~\cite{hu2023animateanyone,guo2023sparsectrl, xu2023magicanimate}.

The primary limitation of most existing video generation models is being restricted to generating very short clips containing limited frames. Inspired by diffusion imputation techniques,
we propose a new autoregressive mechanism that enables us to generate long sequences with high temporal and multi-view consistency.

%% file: secs/03_method.tex
\section{\moniker}
\label{sec:method}

To generate long-range, high-quality \artifacts, we introduce a method that builds on video diffusion models, but adds two crucial ingredients: (1) a method for conditioning on coarse scene layout information, and (2) an autoregressive method for generating video frames that avoids drift---that is, that keeps the imagery on the manifold of natural images---even over long sequences.
We train our system on a novel data source, namely street view imagery and corresponding map information (\Sec{data}).

Our system first trains a diffusion model that jointly generates two frames by iteratively denoising two random noise images. This model also takes as input conditioning information rendered from the given layout for two camera views (Sec.~\ref{sec:system}). Our goal is to generate many consistent frames in our output, however, not just two. For this purpose, we modify the pre-trained two-frame generation model to allow it to operate in an autoregressive \emph{temporal imputation} mode without the need for retraining the model. In this mode, the two random noise images used as input to the model are replaced by noised versions of the frame generated for the current camera view and of the current frame warped into the next camera view, respectively (Sec.~\ref{sec:autoreg}).
With the generated frames, we can also run an optional 3D reconstruction (\Sec{3drecon}) to get a 3D scene model.
We begin this section with a brief review of the basic concepts of diffusion models.

\subsection{Background on Diffusion Models}

The goal of an image generation algorithm is to learn to sample images from an underlying data distribution $p_{\textrm{data}}(x)$. Diffusion models achieve this goal by sequentially denoising samples of random noise $x_{T} \sim \gauss \left(0, \mathbf{I} \right)$ into samples from the data distribution $x_0 \sim p_{\textrm{data}}(x)$. In this context, the (forward) \emph{diffusion process},
\begin{equation}
    p \left(x_t | x_0 \right) = \mathcal{N} \left ( x_t; \sqrt{\alpha_t}x_0, 1 - \alpha_t \mathbf{I} \right),
    \label{eq:noising}
\end{equation}
defines the distribution of $x_t$ as a noised version of $x_0$ at diffusion step $t=0, \ldots, T$; $\alpha_t$ is a predefined noise scheduling term.

The above formulation is approximately inverted by the \emph{reverse diffusion process} 
by iteratively computing
\begin{equation}
    x_{t-1} = \textrm{DDIM}\left(x_t, \epsilon_\theta \left( x_t, t\right), t\right),
    \label{eq:reverse_diffusion}
\end{equation}
where $\epsilon$ is a learned neural network with parameters $\theta$; $\epsilon$ predicts the noise we have added on $x_t$, which is used by a DDIM denoiser \cite{song2020denoising} to compute $x_{t-1}$.

\subsection{Layout-conditioned Scene Generation}
\label{sec:system}
Our objective is to create high-quality \artifacts that adhere to a predefined scene layout and optional text prompts. 
Hence, we design a \artifacts generation system that builds on robust, scalable, and controllable diffusion model architectures. Specifically, we begin by designing a two-view generation module that leverages a pretrained text-to-image diffusion model, incorporates a motion module inspired by AnimateDiff~\cite{guo2023animatediff} to enable two-frame generation, and injects control by integrating a G-buffer--conditioned ControlNet~\cite{zhang2023adding}.

\paragraph{Text-to-Image Foundation Model}
We build our framework on top of a pretrained Latent Diffusion Model~\cite{rombach2022high}.
Given a noise feature map $z_{T}$ along with a text prompt, this model generates a feature map $z_{0}$ by iteratively denoising $z_{T}$ using Eq.~\ref{eq:reverse_diffusion}. The generated feature map is then decoded into a high-quality image $x_{0}$ using decoder $\mathcal{D}, x_{0} = \mathcal{D} (z_{0})$. This decoder, along with a corresponding encoder $\mathcal{E}, x \approx \mathcal{D} \left( \mathcal{E} (x) \right)$, enables the diffusion model to operate in a low-resolution feature space, rather than in image space, for computational efficiency. 
Although not specialized for street-view generation, this foundation model is trained on generic text-to-image generation tasks.
We do not further modify its parameters, and so we naturally inherit text control from this foundation model, even for scenarios that are scarce or absent in our data, e.g., streets full of snow.

\paragraph{Two-Frame Generation via Motion Module}
A critical ingredient in our system is simultaneous generation of two consecutive frames in a \artifact.
We draw inspiration from recent advances in video diffusion models. 
In particular, we follow AnimateDiff~\cite{guo2023animatediff} to insert a motion module into our foundation model.
This motion module enables information exchange across frames at each diffusion step $t$ and effectively extends our noise prediction network to $\epsilon_\theta \left( z_t^{(i)}, z_t^{(i+1)}, t \right)$ and jointly estimates $\left( z_{t-1}^{(i)}, z_{t-1}^{(i+1)} \right)$.
Here, $i$ is the index of the first generated frame and $i+1$ the following. Similar to the single-frame case of Eq.~\ref{eq:reverse_diffusion}, this process starts from random Gaussian noise for both $z_T^{(i)}$ and $z_T^{(i+1)}$. In the following, we refer to this standard video diffusion approach as \textit{\textbf{parallel denoising}}, because two frames are generated in parallel.
The parameters of this motion module are trained with our custom dataset (see Sec.~\ref{sec:data}).

Note that we choose to output two frames as the minimal and most flexible design,
yet one can easily generalize the formulation to more than two frames. That said, current video diffusion models are limited to generating a few tens of frames, so simply adding more than two frames to the parallel denoising process does not scale to the long \artifact sequences we aim to generate.

\paragraph{Layout Conditioning and Camera Control}
While the aforementioned motion module 
can successfully generate two coherent video frames, it does not provide us with direct control over the content. For this purpose, we additionally adopt a ControlNet-based conditioning mechanism~\cite{zhang2023adding}. Specifically, our input street maps and corresponding height maps are rendered from the desired two camera poses $C^{(i)}$ and $C^{(i+1)}$ into G-buffers~\cite{deering1988triangle}, $G^{(i)}$ and $G^{(i + 1)}$. The G-buffers transform our conditioning information from map space into screen space. This conditioning mechanism adds direct control over the camera pose and scene layout and is closely related to depth-conditioned ControlNets used with concurrent video diffusion models \cite{hu2023animateanyone,wang2023motionctrl,xu2023magicanimate}. We illustrate our system in Fig.~\ref{fig:system}.
Note that our system doesn't take camera pose parameters as input.
\input{figs/system}
\input{figs/autoreg_vdif}

\subsection{Autoregressive Video Diffusion}
\label{sec:autoreg}
The parallel denoising method discussed above can generate the first two frames of a \artifact.
For frames beyond those, we would like to ensure that they are consistent with all previously generated frames.
G-buffer conditioning alone cannot guarantee consistency because this approach only controls the layout of the scenes, but not their appearance. 
Thus, stacking pairs of images generated with parallel denoising over time does not achieve the desired consistency, because consistency is only achieved within each pair.
Alternatively, one can also append the generated frame to the ControlNet input, adding RGB control for appearance. However, we empirically find it suffering from severe quality drift.

Inspired by imputation techniques that were used in time-series analysis~\cite{tashiro2021csdi}, image inpainting~\cite{song2020score}, monocular depth estimation~\cite{saxena2023surprising}, and text-conditioned generic video generation~\cite{ho2022video}, we propose a  \textit{\textbf{temporal imputation}} method to enable consistent long-range autoregressive video generation for generating \artifacts. 

Temporal imputation is a surprisingly simple, yet effective idea. 
Importantly, it directly leverages the pre-trained two-frame model without the need to retrain or refine it.
Given the last generated frame $x^{(i)}$, we aim to generate the next frame $x^{(i+1)}$ and continue the same procedure over and over again. 
This is possible using a twist on the reverse diffusion process in \Eq{reverse_diffusion} with pretrained noise predictor $\epsilon_\theta$.
At each reverse diffusion step, instead of reusing the estimation of $x^{(i)}_t$ from the previous diffusion step $t+1$,
we directly sample $x^{(i)}_t$ from the known generated frame $x^{(i)}$,
\begin{equation}
    z_t^{(i)} \sim \gauss \left ( z_t^{(i)}; \sqrt{\alpha_t} \mathcal{E} \left( x^{(i)} \right), 1 - \alpha_t \mathbf{I} \right),
\label{eq:z1}
\end{equation}
Here, Eq.~\ref{eq:z1} essentially adds noise to the latent corresponding to the known frame $x^{(i)}$.
Because we apply \Eq{z1} to all reverse diffusion steps, the ultimate denoised $z^{(i)}_0$ results in the same image $x^{(i)}$ after decoding.
Note that this is only applied to the first frame of the two-frame generation.
The second frame $z^{(i+1)}_t$ still comes from the estimation of step $t+1$.
In our autoregressive setting, this change ensures that the first frame of the next pair of generated images is the same as the last frame of the previous pair. After generating the new pair, however, we discard the first frame of the generated pair and only append the second, $x^{(i+1)}$, to our sequence.

One key insight of our temporal imputation method is that, through imputation, the autoregressive generation takes multiple noised copies of the condition frame $x^{(i)}$ throughout the reverse diffusion process.
This effectively makes the conditioning more robust to any generated slightly out-of-domain frames, because the uncanny details will be submerged by the noise we add.

While the two-frame diffusion model promotes consistency between pairs of generated frames in aforementioned imputation, we empirically observe that the consistency between $x^{(i)}$ and $x^{(i+1)}$ is not always satisfying.
For better consistency, we devise two additional techniques.
First, inspired by image inpainting techniques~\cite{lugmayr2022repaint}, we find that after denoising $z^{(i+1)}_t$ to $z^{(i+1)}_{t-1}$, adding noise to $z^{(i+1)}_{t-1}$ and rerun the imputation can further homogenise  $z^{(i+1)}_{t-1}$ with the noised condition $z^{(i)}_{t}$.
We only proceed to the next step after a few rounds of such back-and-forth.
We refer to this as \emph{Resample}.
Second, we notice that the appearance of $x^{(i+1)}$ is still influenced by the choice of the noisy latent $z_T^{(i+1)}$.
Hence, we propose to warp the known RGB image $x^{(i)}$ from its camera pose $C^{(i)}$ to the pose of the next view $C^{(i+1)}$ using the disparity map available in G-buffer $G^{(i)}$.
Even though the warped image has holes and other artifacts, its noised version provides an excellent choice for $z_T^{(i+1)}$.
We denote this technique as \emph{WarpInit}.
Together, they complete our temporal imputation method (see Fig.~\ref{fig:autoreg_vdif}).
To the best of our knowledge, our system is the first to adopt imputation techniques, specifically, temporal imputation, to autoregressive video generation, using our two-frame diffusion model.

\subsection{\artifact Data}
\label{sec:data}
Because our \artifacts task is unique, we also require a novel dataset that allows us to learn control over both scene layout and camera pose.
A key insight behind our work is to identify geographic data captures from mapping sites like Google Street View~\cite{anguelov2010gsv} as a fitting source of training data for our generative task.
Google Street View offers a large corpus of 
high-quality, diverse, and worldwide street view imagery, geolocated and situated in a geographic context. 
In our work, we show a proof-of-concept use of this data source by collecting $1.5$M images covering 33 km$^2$ from four major cities, Paris, London, Barcelona, and New York City, across four countries.
These data also come with structural information like scene layout and image metadata including the camera pose and capture timestamp. 
This enables our system to associate layout and camera pose control signal with street view generation.
On the other hand, along with its scale, coverage, and multi-modality, come data challenges; in particular, the aerial data is coarse and imperfectly aligned when projected to ground level views.
Thus, our system must be sufficiently robust to these sources of noise.
We show in \Sec{results} that our system generates controlled high-quality \artifacts, a task otherwise impossible without this precious data.

%% file: figs/system.tex
\begin{figure}[t]
\centering
\includegraphics[width=\linewidth]{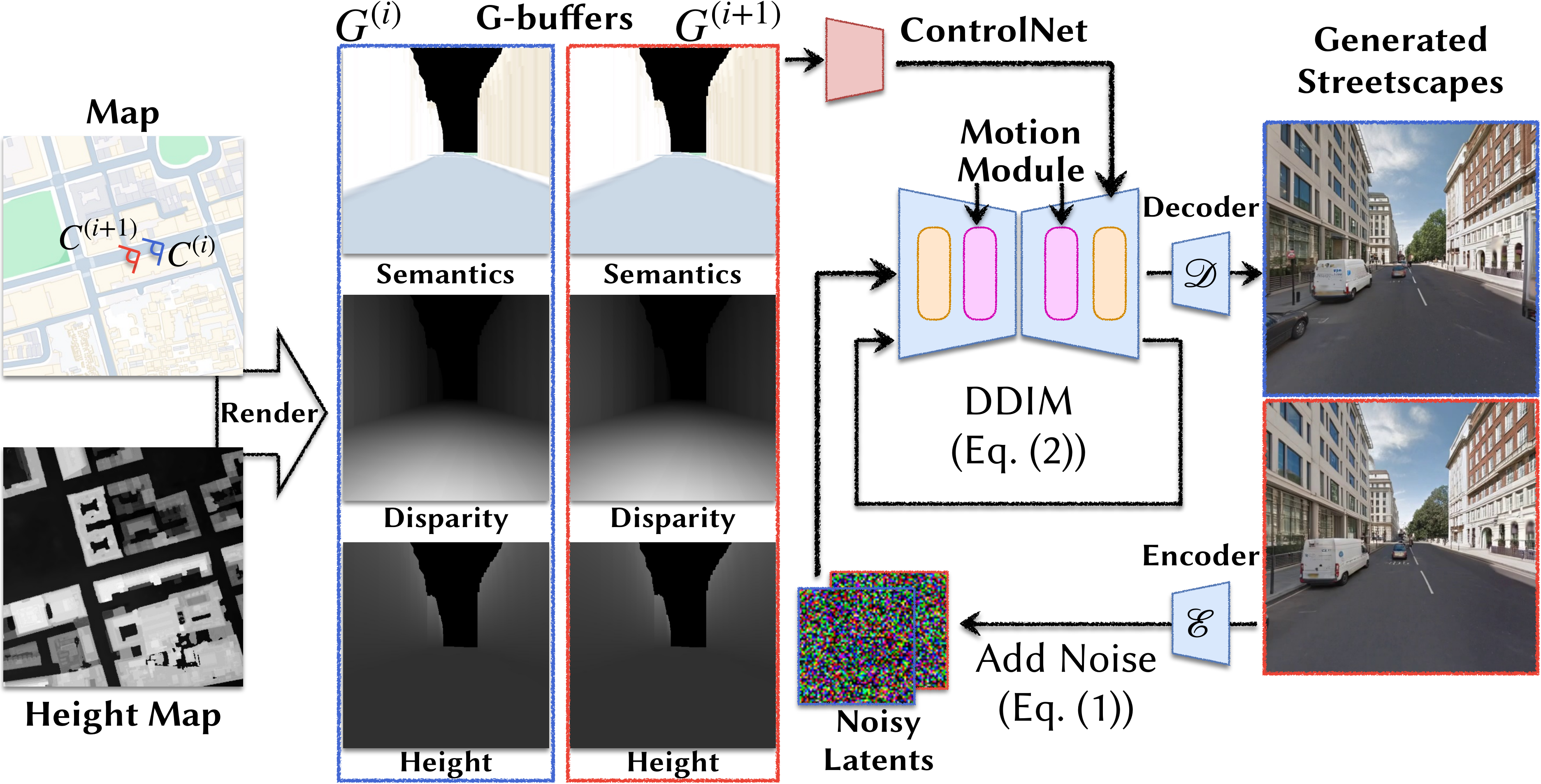}
\caption{
\textbf{Layout-conditioned Scene Generation}.
Using the input scene layout (overhead street map and height map), we render two geometry buffers (G-buffers), $G^{(i)}$ and $G^{(i+1)}$, that contain semantic labels encoded in an RGB image as well as image-space disparity maps and height maps, for camera poses $C^{(i)}$ and $C^{(i+1)}$. These G-buffers condition a motion-aware latent diffusion model that generates a pair of images. Orange and purple boxes
illustrate spatial and temporal layers, respectively.
}
\label{fig:system}
\end{figure}

%% file: figs/autoreg_vdif.tex
\begin{figure}[t]
\centering
\includegraphics[width=\linewidth]{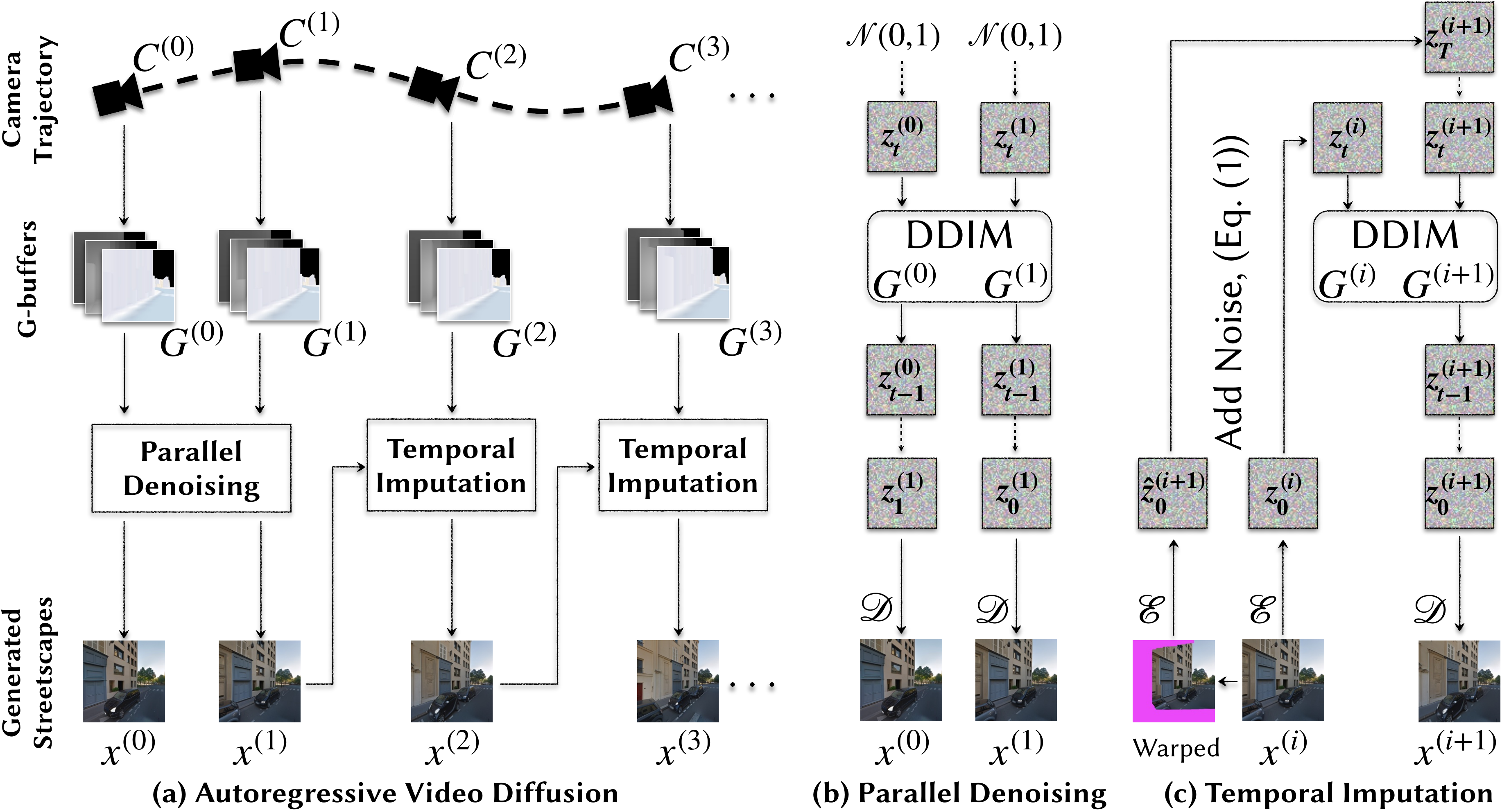}
\caption{
\textbf{Autoregressive Video Diffusion.}
The {\moniker} system generates a sequence of consistent frames along a desired camera trajectory. 
Consistency is achieved by generating the first $2$ frames jointly using \emph{parallel denoising}, then generating each subsequent frame via \emph{temporal imputation}, guided by the previous frame in an autoregressive manner. Both procedures use the same model, but with different reverse diffusion formulations.
}
\label{fig:autoreg_vdif}
\end{figure}

%% file: secs/04_results.tex
\section{Results and Applications}
\label{sec:results}
We validate our \artifacts system on several generative tasks.
We first describe experimental details in \Sec{details}.
In \Sec{streetview_syn}, we present results on a task we call \emph{long-range consistent street view generation}. This task involves sythesizing all images from scratch, given a desired scene layout and camera trajectory.
In \Sec{autoreg_syn} we evaluate on a second task, \emph{perpetual view generation}~\cite{liu2021infinite}, where unlike the first task, we start from a known, real input view,
and autoregressively generate a long sequence of new views. 
In \Sec{application}, we discuss
applications of our \artifacts system, including 
controlling an output \artifact via text description
and mix-and-match map and geographic styles,
as well as an application of interpolating real street view images in ~\Sec{interp_real}.

\subsection{Experimental Details}
\label{sec:details}
\paragraph{Model and Training}
We base our system on a pretrained text-to-image latent diffusion model~\cite{rombach2022high}.
To enable two-frame generation, we insert motion modules as proposed in AnimateDiff~\cite{guo2023animatediff}, but change their $16$-frame formulation to a $2$-frame one.
We further add a ControlNet~\cite{zhang2023adding} to enable layout and camera pose control.
In contrast to the standard single-image ControlNet, we form a two-frame ControlNet by also cloning the motion modules.
Hence, our ControlNet contains both spatial and temporal modules.
We use the diffusion noise prediction objective from DDPM~\cite{ho2020denoising} to train our model with learning rate of $1e^{-5}$ and batch size $256$ for $1.5$M iterations.
Note we keep the base diffusion model U-Net intact in training, so as to preserve the text control from the base model.
Our model is trained using image data from Google Street View.\footnote{Google Maps Street View images used with permission from Google.}
For two-frame training, we group pairs of $512\times512$ images captured within $80$s and $5$ meters.

\paragraph{Inference}
We use both parallel sampling and temporal imputation in our generations, as defined in \Sec{method}.
For parallel sampling, we follow the deterministic sampling process in DDIM~\cite{song2020denoising}.
We use $50$ denoise steps with the classifier-free guidance~\cite{ho2021classifier} scale set to $7.5$.
For the temporal imputation stage defined in \Sec{autoreg}, we use $10$ Resample steps and WarpInit.
Note for any inference in our experiments, we use maps from regions not seen during training.

\subsection{Long-range Consistent Street View Generation}
\label{sec:streetview_syn}
\input{figs/streetview_syn}
We first evaluate our system on the task of generating long-range consistent street views of a large-scale urban scene, where we \emph{generate all views} in the output from scratch.
In this task, the model is given a street map, its corresponding height map, and a sequence of camera poses defining a camera trajectory, and generates consistent street views corresponding to the trajectory.
We compare our 
method 
to InfiniCity~\cite{lin2023infinicity}, a state-of-the-art consistent street view generation system.
Unlike \moniker, 
InfiniCity requires aligned CAD models as training data, and 
it only works for cities where CAD models are readily available data (just London, in their case).
For a fair comparison, we also only use maps from London when evaluating our system.
We show example results in \Fig{streetview_syn}.
We find that \artifacts produced by our system, either the $2$-frame version or the $4$-frame version, are consistently more realistic compared to those of InfiniCity.
Our results have higher-fidelity details such as windows, pavement, and vegetation.
In addition, our results simulate richer lighting effects, including natural shadows (missing in InfiniCity's results).
To quantitatively evaluate the generated results, we use FID and KID metrics. These metrics measure the distance between distributions defined by the generated results and by ground truth imagery, and are commonly used to evaluate generative models.
We summarize the results in \Tab{streetview_syn}.
While our system generates \artifacts autoregressively, we observe in \Tab{streetview_syn} that our quality is consistently better than InfiniCity's across all autoregressive steps by a fair margin.
All InfiniCity results are obtained from the authors.

In addition to our comparison to InfiniCity, in \Fig{long_track}, we also show two example \artifacts from our system, simulating long-range, synthesized street cruises.
Each scene spans a map of $\sim$ 1~km$^2$.
We generate $100$ frames for each scene with flexible camera motion,
to mimic ``looking around'' the synthesized city.
We find that our system robustly generates high-quality street views across long camera trajectories with various camera motions.
\input{tbls/streetview_syn}

\subsection{Perpetual Street View Generation}
\label{sec:autoreg_syn}
\input{tbls/autoreg_syn}
Core to our system is our novel autoregressive video diffusion method.
To further validate this method, we also compare it with alternative autoregressive view generation methods.
We choose the perpetual view generation~\cite{liu2021infinite} task, tailored to street scenes, as the benchmark.
For this task, we take a real street view image as a starting view, and then generate new views autoregressively along a desired camera trajectory, with the goal of generating views that are consistent with the starting image.
We additional provide our system with the corresponding scene layout for the starting view.
We compare to several state-of-the-art
methods:
\begin{itemize}
    \item \emph{InfiniteNature-Zero (InfNat0)~\cite{li2022_infinite_nature_zero}} is a GAN-based method. It uses scene geometry estimated via monocular depth prediction to forward-warp views during generation.
    We test two variants of InfNat0, one using the original monocular depth formulation (\emph{InfNat0-mono}) and one using proxy geometry derived from the input height map (\emph{InfNat0-proxy}).
    \item \emph{DiffDreamer~\cite{cai2022diffdreamer}} is a diffusion-based alternative to InfNat0-proxy that replaces the GAN approach with a more powerful diffusion model.
    \item \emph{Zero123~\cite{liu2023zero}} is a diffusion-based generative method that does not use explicit 3D geometry.
    It takes in an input view and a relative camera pose 
    directly as condition embedding parameters. Note that the original Zero123 always conditions on the given first view because it focuses on object generation, which has a canonical, bounded camera space. We modify Zero123 to generate unbounded long-range street views autoregressively by conditioning on the last generated view. We refer this variant as \emph{Zero123-A}.
\end{itemize}
We obtain results for InfNat0-mono and InfNat0-proxy using the code from its authors to train the model on our data.
For diffusion-based approaches (DiffDreamer and Zero123-A), for fair comparison, we train these models using the same pretrained diffusion base models as on our data.

We show examples of generated views in \Fig{autoreg_syn}.
We find that at the first autoregressive step, most methods generate high-quality views.
At step-$20$, the quality of DiffDreamer's results start to decrease.
At step-$40$, the realism of samples from InfNat0-mono and DiffDreamer have deteriorated, and the results from Zero123-A and InfNat0-proxy are also quite poor.
In contrast, our \artifacts maintain consistent, realistic quality across all steps.

For quantitative evaluation, we follow Infinite Nature~\cite{liu2021infinite} and evaluate with two kinds of metrics: First, we compute \emph{short-range accuracy} for the first few images in the sequence, where we have known ground truth street view imagery, and where the view synthesis problem has not yet become fully generative due to overlap with the seed view.
For this metric, we compute LPIPS~\cite{zhang2018perceptual} to measure the holistic image similarity over the first $16$ views.
We prefer the holistic measure LPIPS over local per-pixel errors because even for early views, the task still has generative aspects due to occlusions, super-resolution from zooming-in, and scene dynamics such as moving cars.
Second, we evaluate \emph{long-range image quality} for more distant views (where the task is almost entirely generative), we compute FID and KID between generated street views and real street views.
We show quantitative comparisons in \Tab{autoreg_syn}.
From the comparison, we conclude that our generated \artifacts have short-range accuracy comparable to a state-of-the-art method, Zero123-A.
Meanwhile, unlike prior methods that suffer from serious degradation in long-range quality (e.g., after $32$ steps), our method scales well such long sequences.

\subsection{Creative Applications of {\moniker}}
\label{sec:application}
Beyond standard view generation tasks, our \artifacts system enables numerous creative applications.
We showcase two example use cases of our system: text-controlled \artifacts generation and geographic style mix-and-match.

\paragraph{Text Controlled \artifacts}
Thanks to the use of a pre-trained text-to-image model as the foundation of our system, we can control our \artifacts generation using text prompt.
While our dataset mainly contains street view images in the daytime without extreme weather conditions, due to the real-world street view capture constraints, our generation can hallucinate various conditions on the street, including both weather and time-of-day.
For instance, we can ask for ``Paris, Before Sunset'' or ``A Rainy Day in New York''.
In \Fig{text_control}, we display a few example text-controlled \artifacts with specific weather and time-of-day.
We find that our system can effectively generate \artifacts reflecting the given text description.

\paragraph{What Makes Paris Look Like New York?}
In a seminal prior work, researchers investigated ``What Makes Paris Look Like Paris?''~\cite{doersch2012makes}.
Our \artifacts system imagines what a street looks like from a map.
Therefore, in this work, we make an alternative quest to our system --- \emph{What makes Paris look like New York}?
Here, ``Paris'' denotes the scene layout --- we take a street map of Paris blocks; ``New York'' indicates the geographic style of the generation --- we hope our generated \artifacts look distinctly like New York City.
In \Fig{loc_mixmatch}, we show our generated \artifacts for this task.
We observe that our system can mix-and-match the layout and the geographic style and generate \artifacts that demonstrate distinguishable \emph{signature} details of the target location, e.g., New York style street sign post and Parisian Juliet balcony (or ``balconet'').

%% file: figs/streetview_syn.tex
\begin{figure*}[ht]
\centering
\includegraphics[width=0.95\linewidth]{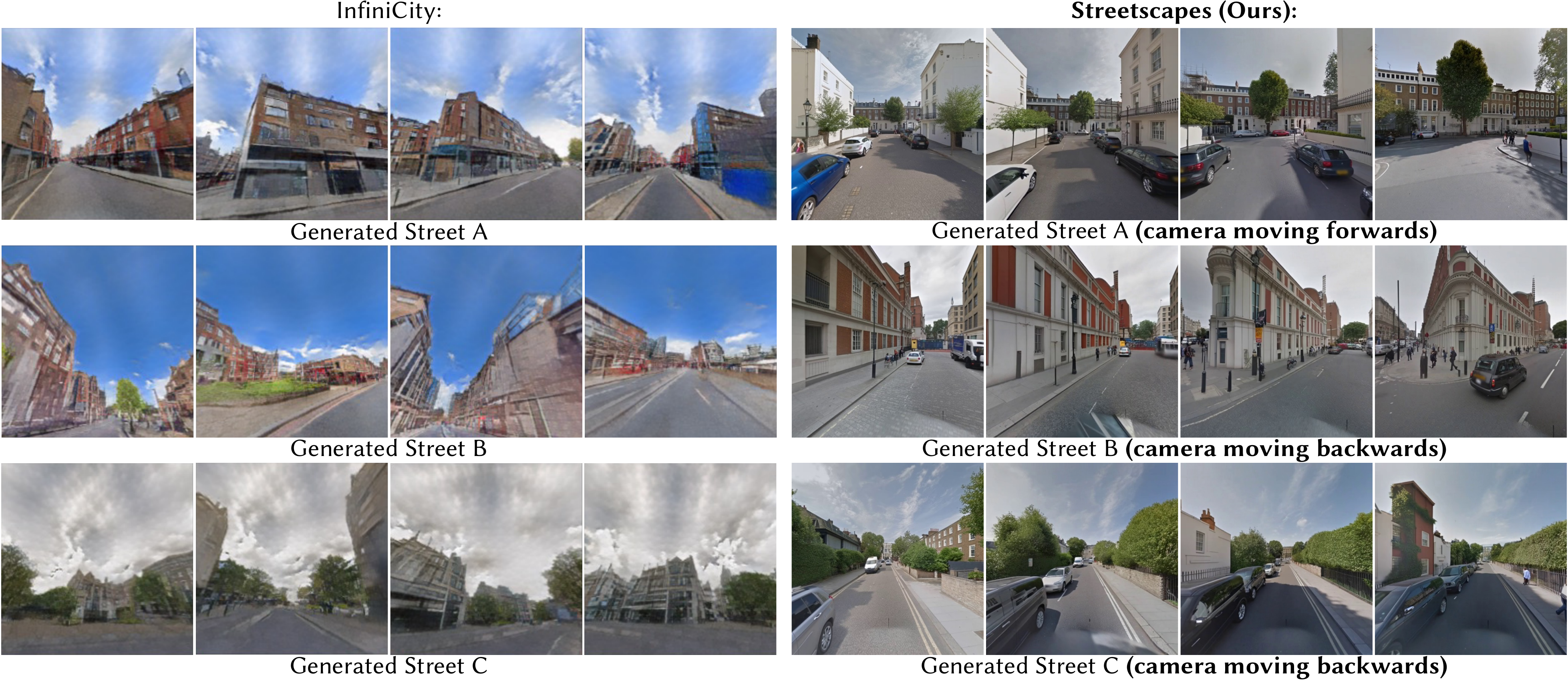}
\caption{
\textbf{Long-range Consistent Street View Generation}.
We compare {\moniker} to a state-of-the-art street view generation approach, InfiniCity~\cite{lin2023infinicity}, on the task of generating consistent views on long paths through large-scale street scenes.
In this task, we let {\moniker} generate consistent views autoregressively following various camera tracks.
We find that {\moniker} consistently generates higher-fidelity street views that are noticeably more realistic than InfiniCity's results.
}
\label{fig:streetview_syn}
\end{figure*}

%% file: tbls/streetview_syn.tex
\begin{table}
\caption{
\textbf{Large-scale Consistent Street View Generation.}
We compare \moniker to InfiniCity on image quality metrics FID and KID.
Because we generate \artifacts autoregressively, we evaluate \artifacts quality at different autoregressive steps.
We find that our quality is significantly better than InfiniCity's, across all steps.
}
\label{tbl:streetview_syn}
\begin{center}
\resizebox{0.8\linewidth}{!}{
\begin{tabular}{lccccc} 
\toprule
\multirow{2}{*}{Metric} & \multirow{2}{*}{InfiniCity} & \multicolumn{4}{c}{\artifacts @ Steps}\\
& & all & 1--16 & 16--32 & 32--64\\
\midrule
FID$\downarrow$ & $108.47$ & $17.79$ & $16.12$ & $21.70$ & $29.93$ \\
KID$\downarrow$ & $0.084$ & $0.023$ & $0.017$ & $0.022$ &  $0.025$ \\
\bottomrule

\end{tabular}
} 
\end{center}
\end{table}

%% file: tbls/autoreg_syn.tex
\begin{table}
\caption{
\textbf{Perpetual Street View Generation.}
We compare our \artifacts with results from state-of-the-art autoregressive view generation methods
(\colorbox{tbl_no1}{best}, \colorbox{tbl_no2}{$2_\text{nd}$ best}).
We compute a holistic image similarity measure (LPIPS) to assess consistency in the near-range (1--16 steps).
For image quality, we measure the FID and KID at different autoregressive step ranges.
These results indicate that \artifacts have similar near-range quality and consistency to Zero123-A, while maintaining image quality over significantly longer ranges ($32+$ steps) than all alternative methods.
}
\label{tbl:autoreg_syn}
\begin{center}
\resizebox{\linewidth}{!}{
\begin{tabular}{lcccc} 
\toprule
\multirow{2}{*}{Metric} &  LPIPS $\downarrow$ & \multicolumn{3}{c}{FID$\downarrow$/KID$\downarrow$ @ Steps}\\
 & 1--16 & 1--16 & 16--32 & 32--64 \\
\midrule
InfNat0-mono & $0.600$ & $16.72$/$0.014$ & $48.97$/$0.042$ & $71.10$/$0.056$ \\
InfNat0-proxy & $0.614$ & $26.80$/$0.025$ & $51.48$/$0.052$ & \cellcolor{tbl_no2}{$59.00$/$0.048$} \\
DiffDreamer & $0.589$ & $45.96$/$0.050$ & $164.23$/$0.208$ & $207.37$/$0.242$ \\
Zero123-A & \cellcolor{tbl_no1}{$0.497$} & \cellcolor{tbl_no1}{$7.09$/$0.005$} & \cellcolor{tbl_no2}{$34.64$/$0.035$} & $64.77$/$0.062$ \\
\midrule
\artifacts & \cellcolor{tbl_no2}{$0.519$} & \cellcolor{tbl_no2}{$11.69$/$0.012$} & \cellcolor{tbl_no1}{$25.59$/$0.028$} & \cellcolor{tbl_no1}{$35.47$/$0.034$} \\
\bottomrule

\end{tabular}
} 
\end{center}
\end{table}

%% file: secs/05_conclusion.tex
\section{Discussion and Conclusions}
\label{sec:discussion}

\paragraph{Limitations}
While our system generally generates high-fidelity, realistic \artifacts, we observe occasional unnatural movements of transient objects in our \artifacts, e.g., a car suddenly appearing or disappearing.
We hypothesize the cause to be the limited capture frequency of street view data, and the pervasive presence of such objects in the data.
Explicitly modeling transient objects and  allowing for control of them would be a promising area of exploration.
In addition, our results exhibit some short-range inconsistencies (e.g., slight color flicker from frame to frame), due to a balance between quality and consistency that is inherent to many diffusion problems.

\paragraph{Conclusion} We have designed a system called \artifacts capable of generating highly realistic, consistent street views of a synthesised urban scene. 
The generated imagery can be controlled by scene layout, camera pose, and optionally, a ``style'' text input.
This system is only possible due to our choice of the scale and high quality two-frame diffusion model, as well as identifying street view and map corpora as a unique data source.
Furthermore, we overcome the limitation of video diffusion models to a finite number of frames via an autoregressive video diffusion method powered by our novel temporal imputation approach, enabling truly long-range view generation.
Our work marks a forward step in visual generative models, enriching generative capabilities from objects to unbounded large-scale scenes.

\begin{acks}
We thank Thomas Funkhouser, Kyle Genova, Andrew Liu, Lucy Chai, David Salesin, David Fleet, Jonathon T. Barron, Qianqian Wang, Shiry Ginosar, Luming Tang, Hansheng Chen, and Guandao Yang for their comments and for constructive discussions.
We thank William T. Freeman and John Quintero for helping review our draft.
We thank all anonymous reviewers for their helpful suggestions.
G.W. was in part supported by Google, Samsung, and Stanford HAI. B.D. was supported by a Meta PhD Research Fellowship.
The initial idea of this project was partly inspired by \href{https://youtu.be/0S43IwBF0uM?si=CrKO-ui4LX03xRjk}{the Star Guitar video} created by Michel Gondry and The Chemical Brothers.
\end{acks}

%% file: figs/long_track.tex
\begin{figure*}[ht]
\centering
\includegraphics[width=0.95\linewidth]{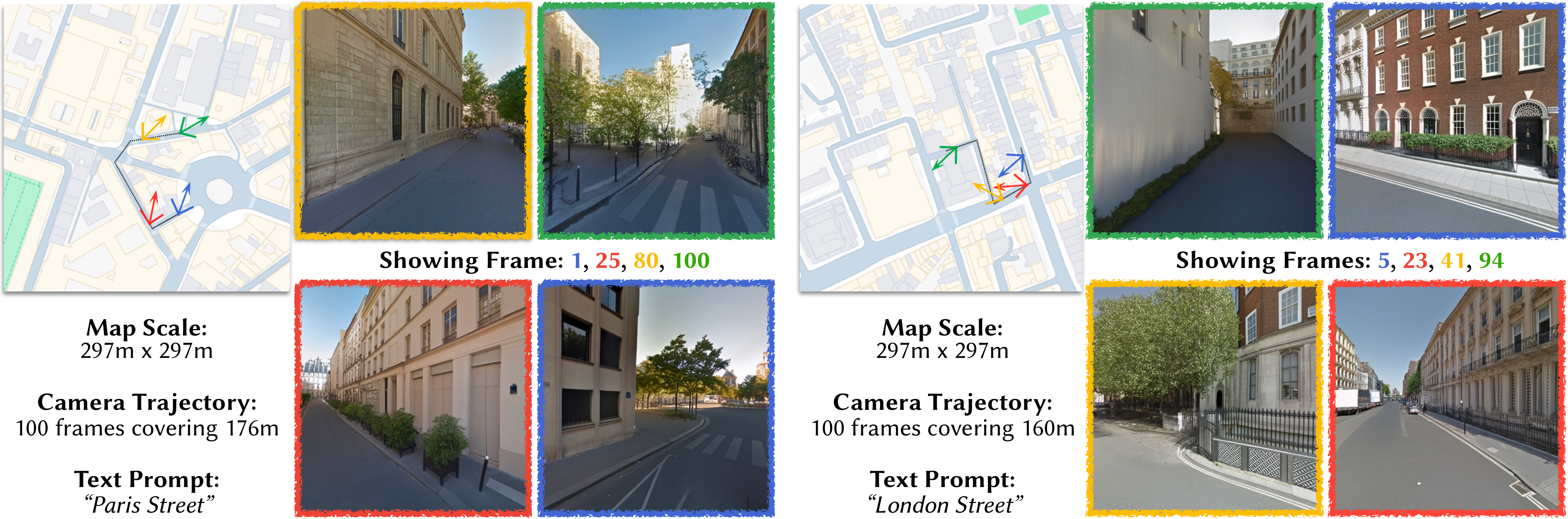}
\caption{
\textbf{Long-range Street Cruise}.
While alternative autoregressive generation methods suffer from severe quality degradation after $32$ frames (see \Fig{autoreg_syn} and \Tab{autoreg_syn}), our \artifacts system can generate 
long street cruises without noticeable drift in quality for $100$ frames along paths of over $170$ meters spanning multiple city blocks.
Note how our results are consistent with the specified scene layout and camera poses (illustrated on the maps).
In addition, in contrast to prior methods that support either forward-only~\cite{liu2021infinite} or backward-only~\cite{SceneScape} camera motion, our system allows for flexible camera control wherein the user can freely move and turn the camera.
}
\label{fig:long_track}
\end{figure*}

%% file: figs/autoreg_syn.tex
\begin{figure*}[ht]
\centering
\includegraphics[width=.95\linewidth]{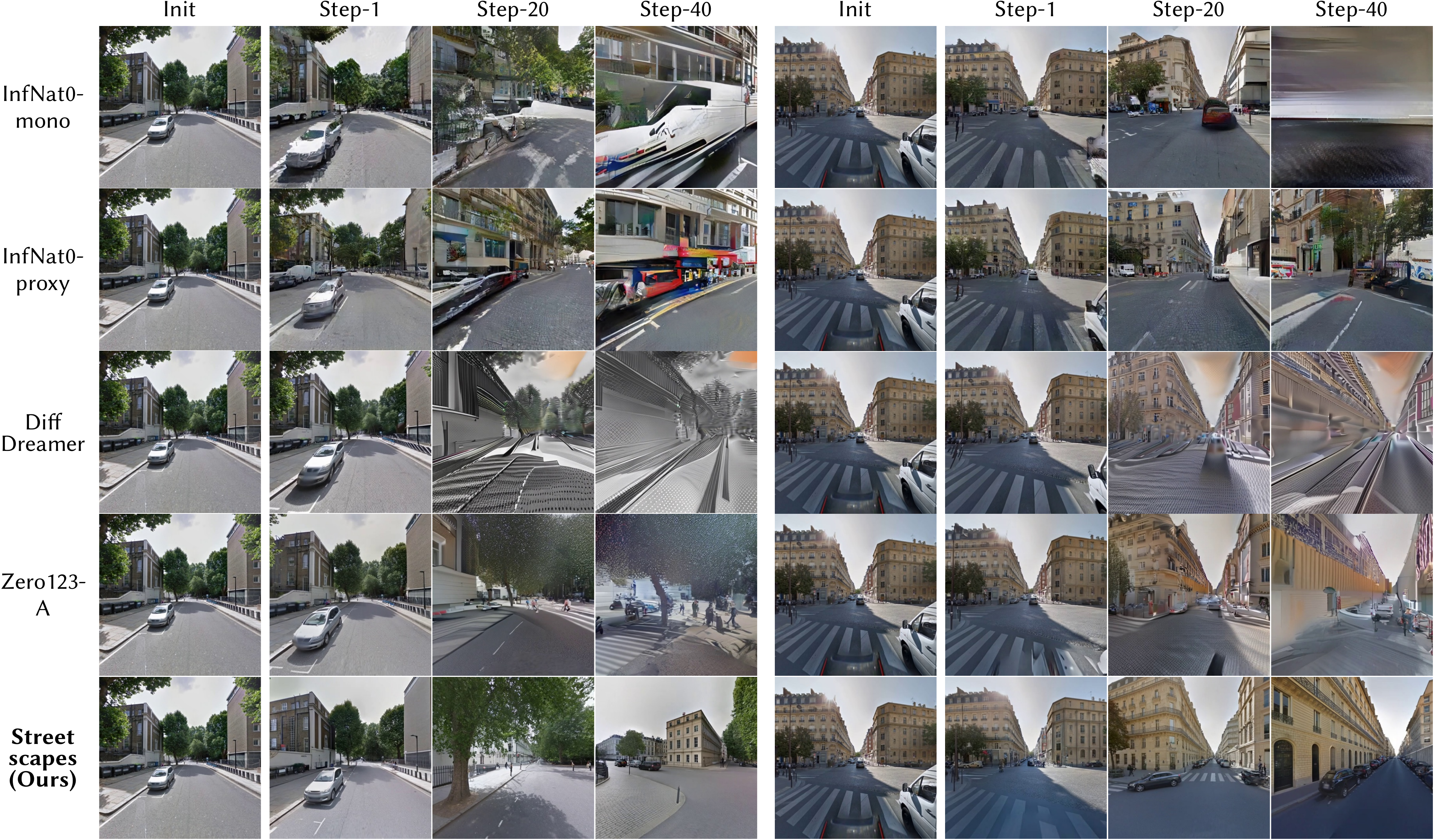}
\caption{
\textbf{Perpetual Street View Generation}.
We compare {\moniker} with state-of-the-art autoregressive view generation methods, including InfiniteNature-Zero~\cite{li2022_infinite_nature_zero} (using either monocular depth, \ie InfNat0-mono, or proxy depth, \ie InfNat0-proxy), DiffDreamer~\cite{cai2022diffdreamer}, and an autoregressive variant of Zero123~\cite{liu2023zero}, \ie Zero123-A.
In this task, we are given an initial input image and a camera track.
Each method aims to generate a consistent street view following the camera track autoregressively.
We pick generation step-$1$ to demonstrate degradation-free generation quality, as well as step-$20$ and step-$40$ for long-range generation quality.
Note how the results of our {\moniker} method remain highly realistic, while those of other methods degrade significantly.
}
\label{fig:autoreg_syn}
\end{figure*}

%% file: figs/text_control.tex
\begin{figure*}[ht]
\centering
\includegraphics[width=0.98\linewidth]{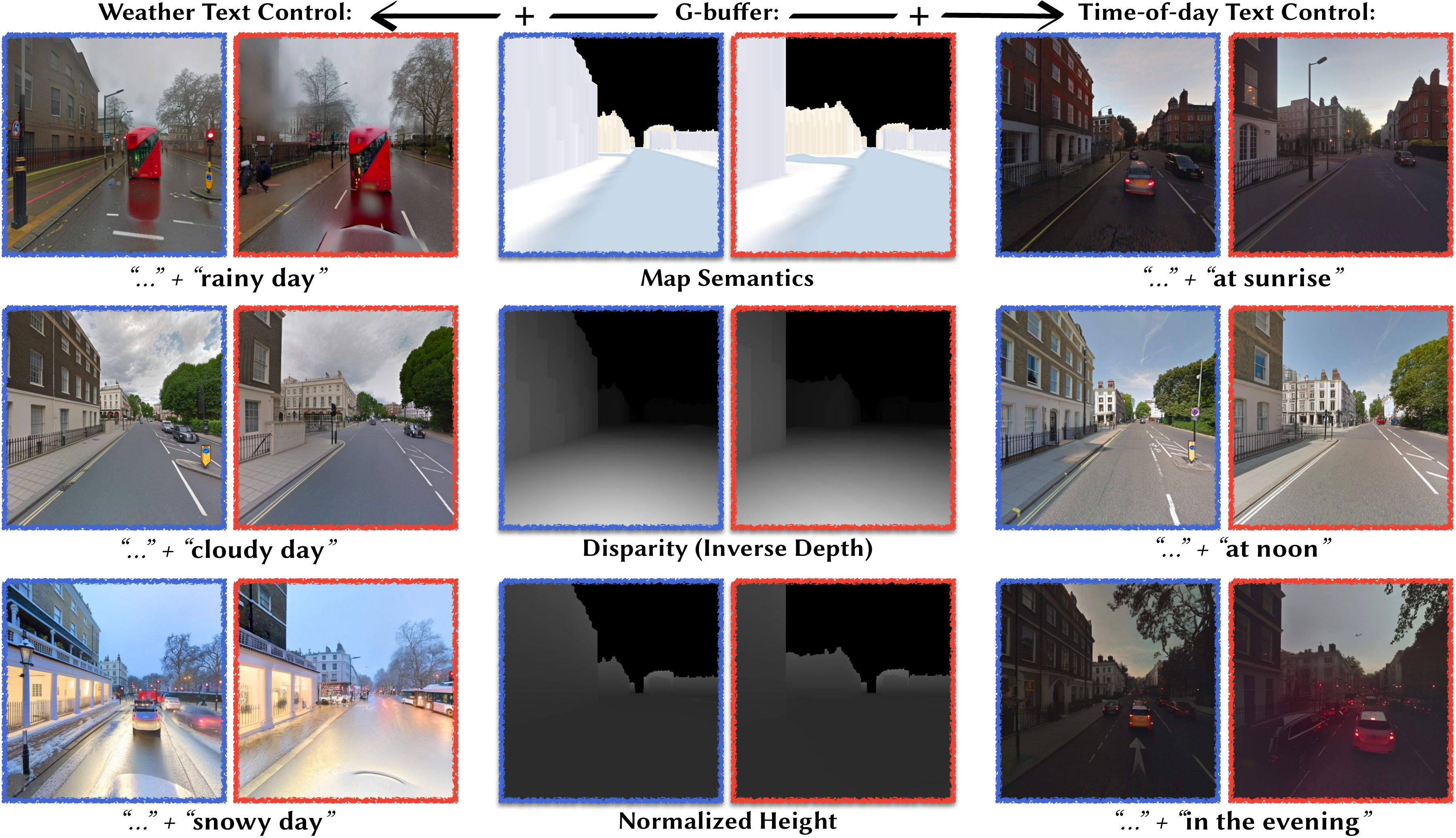}
\caption{
\textbf{Text-Controlled \artifacts}.
We build our \artifacts system on a pre-trained text-to-image model.
This enables text-prompt based interaction with the generation process.
We demonstrate a few examples of text-controlled \artifacts in this figure.
Given the same map and camera poses (and hence the same G-buffers), our system can condition on time of day and
weather conditions described by text prompt.
It generates high-quality street views in diverse conditions consistent with the text description.
}
\label{fig:text_control}
\end{figure*}

%% file: figs/loc_mixmatch.tex
\begin{figure*}[ht]
\centering
\includegraphics[width=0.98\linewidth]{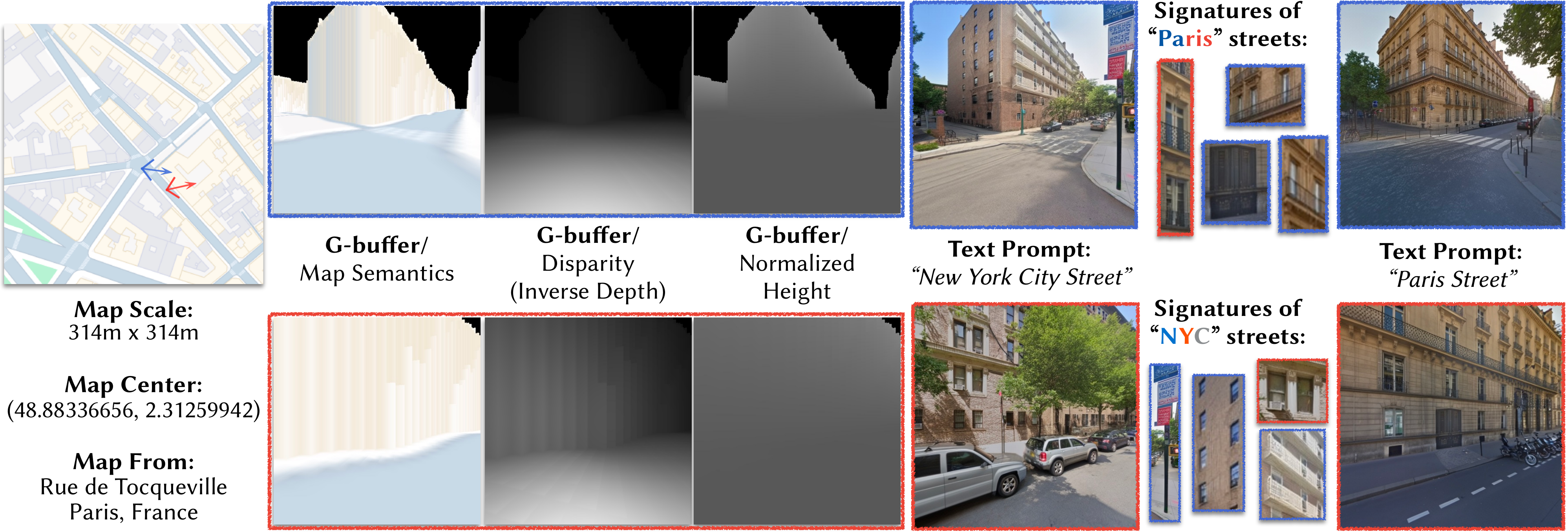}
\caption{
\textbf{What Makes Paris Look Like New York?}
We demonstrate an example of mix-and-match scene layout with geographic styles using our \artifacts system.
Given a street map from Paris, we can generate \artifacts that look like New York City, thanks to our disentanglement of layout and scene appearance, and to the text control inherited from text-to-image models.
Following prior works~\cite{doersch2012makes}, we could analyze our generation to summarize the ``signature'' of a geographic style our system learns, i.e., ``what makes Paris look like New York''? For instance, in the above example, our model learns the balustrade window and French balconies unique to Paris, and the very NYC-style road sign posts.
}
\label{fig:loc_mixmatch}
\end{figure*}

%% file: supplement.tex
\clearpage
\appendix

\section{Extension to $4$-frame Video Diffusion Models}
In \Sec{method}, we discuss the \artifacts system based on a $2$-frame video generation model.
The same system design also applies to models that generate more frames at once. 
We validate this generalization by extending the video generation model to a $4$-frame version. In a $4$-frame model, the Parallel Denoising step in \Fig{autoreg_vdif} generates four frames, and the Temporal Imputation step uses the last two generated frames as the conditioning to generate the next two frames.
We find in our experiments that the $4$-frame model improves the consistency, likely due to increased context size.
On our \href{https://boyangdeng.com/streetscapes/}{\color{magenta}{project page}}, we show result videos from this  $4$-frame model.

\section{3D Reconstruction from \artifacts}
\label{sec:3drecon}
The aforementioned system generates \artifacts that are reasonably multi-view consistent.
Nevertheless, for applications such as VR/AR that demands perfect 3D consistency, generating videos alone may not meet the criteria.
Hence, we introduce an optional 3D reconstruction step to our \artifacts system that reconstructs a Neural Radiance Field~\cite{mildenhall2020nerf} from generated \artifacts.
We base our reconstruction on Zip-NeRF~\cite{barron2023zipnerf}.
Similar to prior works~\cite{wu2023reconfusion, gao2024cat3d}, we add an LPIPS~\cite{zhang2018perceptual} loss to steer the focus of reconstruction to holistic rendering quality instead of pixel-level details that might not be perfectly consistent in \artifacts.
Additionally, we find that using Temporal Imputation with SDEdit~\cite{meng2021sdedit} to resample each frame conditioning on neighbouring frames and picking the sample that is closest to the mode helps remove random transient objects and stabilizes the results.
Please see 3D reconstruction results on our \href{https://boyangdeng.com/streetscapes/}{\color{magenta}{project page}}.

\section{Interpolating Sparse Street View Images}
\label{sec:interp_real}
In addition to generating non-existing realistic streets, \artifacts can also assist the virtual exploration of real streets, thanks to the flexible conditioning of Temporal Imputation.
In particular, we can interpolate real street view captures, which by themselves are too sparse to simulate smooth experiences of driving or walking along a street.
Instead of conditioning on the first two frames in Temporal Imputation using a $4$-frame model, we can condition on the first and last frame and use Temporal Imputation to generate the two frames in between.
This effectively interpolates between the sparse image sequence, increasing the frame rate by about $3$x.
Applying such interpolation iteratively can further increase the frame rate.
Alternatively, we can run the 3D reconstruction described in \Sec{3drecon} to obtain a 3D model, from which we can re-render new paths at any camera speed.
Please see our interpolation results on our \href{https://boyangdeng.com/streetscapes/}{\color{magenta}{project page}}.

\section{Challenges with Geographic Data}
\label{sec:data_issue}
\input{figs/data_error}
Some characteristics of the map and street view image data we train from lead to robustness challenges for our \artifacts system due to data noise when used at scale without manual curation.
We observe two main types of noise in our data:
1) The aerial height map data we use can be coarse and noisy, due to aerial capture precision constraints.
This results in misalignment between our G-buffers and the actual street view images. We show examples of such misalignment in \Fig{data_error}.
2) The camera poses are not always pixel-accurate. To obtain camera poses, we use the latitude and longitude from the data source as coordinates on the ground plane, though we estimate the height above ground ourselves, using the approximate height of the capture vehicle as a proxy estimate.
As a result of approximations in the calculated pose, our G-buffers deviate from perfectly aligned G-buffers.
To mitigate the effects of this noise, we design our system to be more driven by images, via the choice of training two-frame motion modules directly from street view image pairs.
Additionally, our choice of ControlNet for layout and camera control is robust to infrequent noise in the control signal in training, which helps ensure our generation quality after training even with the presence of imperfect G-buffers.
In addition to these issues, some source image regions are blurred for privacy reasons.
We use street view metadata to filter out these examples.
For instance, we remove images with blurred areas over $0.04\%$.

\section{Ablation Study on Autoregressive Video Diffusion}
\label{sec:ablation}
\input{tbls/ablation}
\input{figs/ablation_rgbcontrol}
\input{figs/ablation_resample_warpinit}
In Secs.\ 4.2 and 4.3 of the main paper, we compared our diffusion model--based \artifacts generation system to GAN-based approaches such as InfiniCity and InfNat0, and showed that diffusion models lead to higher generation quality.
In this section, we conduct ablation studies on the design choices of our proposed autoregressive video diffusion method.
Specifically, we investigate:
\begin{enumerate}
  \item the choice of imputation-based autoregressive generation against an alternative autoregressive approach using our system by adding an extra RGB control channel to the ControlNet, where for the condition frame it is the condition image and for the target frame it is all black (\emph{RGB Control}).
  \item the impact of resampling in temporal imputation (Ours vs Ours - Resample).
  \item the impact of initialize each temporal imputation from a noised image warped from the condition image to the target view (Ours vs Ours - WarpInit).
\end{enumerate}
We compare all variants on the perpetual street view generation task (Sec. 4.3) where we can compare both near-range accuracy and long-range image quality.

From the results in \Tab{ablation}, we first find that the alternative RGB Control approach still suffers from severe quality degeneration.
This again supports our key insight that using multiple noised copies of the condition image in imputation-based autoregressive generation is much more robust than simply conditioning on the image itself as input.
In \Fig{ablation_rgbcontrol}, we show the significant difference in image quality between RGB Control and temporal imputation after just $20$ steps in autoregressive generation.

While imputation generally improves image quality over long-range autoregressive generation, as shown in \Tab{ablation} for all variants of temporal imputation, we aim to strike a balance between near-range accuracy (\ie alignment with the given view) and long-range quality.
Inspired by prior works~\cite{lugmayr2022repaint, meng2021sdedit, bansal2024universal} that apply diffusion models to image processing problems, we adopt Resample and WarpInit to our imputation-based autoregressive generation pipeline.
Resample is a common practice to homogenize diffusion generation when partial guidance signals are injected~\cite{lugmayr2022repaint, bansal2024universal}. It can be connected to SDE solvers for the stochastic sampling in the diffusion denoising process.
WarpInit is also a principled guidance approach adopted from SDEdit~\cite{meng2021sdedit} in image editing.
We find in \Tab{ablation} that both Resample and WarpInit help increase the similarity of generated \artifacts to the reference images.
This is further backed by our qualitative results in \Fig{ablation_resample_warpinit}, where we find that among a few random samples from each temporal imputation variants, our full method with both Resample and WarpInit generates the most similar images to the reference, indicating the best near-range accuracy.

\section{More Training Details}
\label{sec:more_details}
We base our system on a pretrained text-to-image latent diffusion model~\cite{rombach2022high}.
To enable two-frame generation, we first insert motion modules as proposed in AnimateDiff~\cite{guo2023animatediff}, but change their $16$-frame formulation to a $2$- or $4$-frame one.
We conduct a two-stage training procedure for our diffusion model.
First, we train \emph{only} the motion modules but keep parameters of the base diffusion model U-Net frozen.
We also use per-location text in our training.
For each image, we 
associate it with the text 
``[location] street'' where ``[location]'' is the city name for the image, e.g., ``London''.
After training the motion modules, we train a ControlNet~\cite{zhang2023adding} for layout and camera pose control.
while training the ControlNet, we keep both the base diffusion model and the base motion modules frozen.
For both training stages, we use the standard diffusion noise prediction objective from DDPM~\cite{ho2020denoising}.
We train the motion modules with a learning rate of $1e^{-5}$ and batch size at $256$ for $1$M iterations.
We train the ControlNet with a learning rate of $1e^{-5}$ and batch size of $256$ for $500$k iterations.

\section{Video Diffusion Models vs 3D Reasoning}
In our system, we choose to use Video Diffusion Models (VDMs) trained on street view image sequences to generate consistent \moniker, instead of implementing exact 3D rendering in our generated views.
This design choice is driven by two observations. 
The first is the underlying data noise. 
As described in \Sec{data_issue}, street view data is noisy, with issues including inconsistency across modalities (e.g. map-vs-image), inaccurate and coarse height maps, frequent transient objects, and inaccurate camera poses. This noise is a significant challenge to generative methods that rely on exact 3D reasoning, as shown by the results of Infinite-Nature and DiffDreamer in Table 2 of our main paper. Second, VDMs are becoming more and more consistent (as evidenced by methods like Sora~\cite{videoworldsimulators2024}), and seem like a promising route towards generating large-scale 3D scenes. Yet they are difficult to control and still can't scale to full city-scale scenes. Hence, in our \moniker system, we addresses challenges of controllability and scaling via G-buffer conditioning and autoregressive generation.

\section{Recent Progress in Video Diffusion Models}
As we were developing our \artifacts system using the best Video Diffusion Model design that was publicly available, namely AnimateDiff~\cite{guo2023animatediff}, there are exciting concurrent progresses in video diffusion models.
For example, WALT~\cite{gupta2023photorealistic} uses a Diffusion Transformer architecture to enable longer and more stable video generation.
\cite{bar2024lumiere} changes the latent diffusion to a pixel diffusion framework and uses multiple levels of spatiotemporal compression and decompression to improve the duration and quality of its video generations.
Thanks to our system design, both methods are compatible with our \artifacts system. 
Indeed, incorporating these models into \artifacts would be an interesting future direction.
Even more impressive is the concurrently released Sora~\cite{videoworldsimulators2024}, which already ``generates videos with stunning geometrical consistency''~\cite{li2024sora}.
This provides further evidence that our design choice of building \artifacts systems based on video diffusion models is promising.

\section{\artifacts Beyond Street View}
In this work, we aim to design a globally applicable system.
We identify Google Street View as our data source that can potentially provide global coverage.
Therefore, we only conduct experiments on street view data.
However, our system design, in part or whole, is applicable to other datasets such as indoor datasets with floorplans (leading to building walk-through generations).
Also, our autoregressive video generation approach is applicable to any diffusion-based video generation model.
It would be an interesting follow-up direction to study incorporating this technique to recent video diffusion models such as WALT~\cite{gupta2023photorealistic} and Lumiere~\cite{bar2024lumiere}.

%% file: figs/data_error.tex
\begin{figure}[t]
\centering
\includegraphics[width=\linewidth]{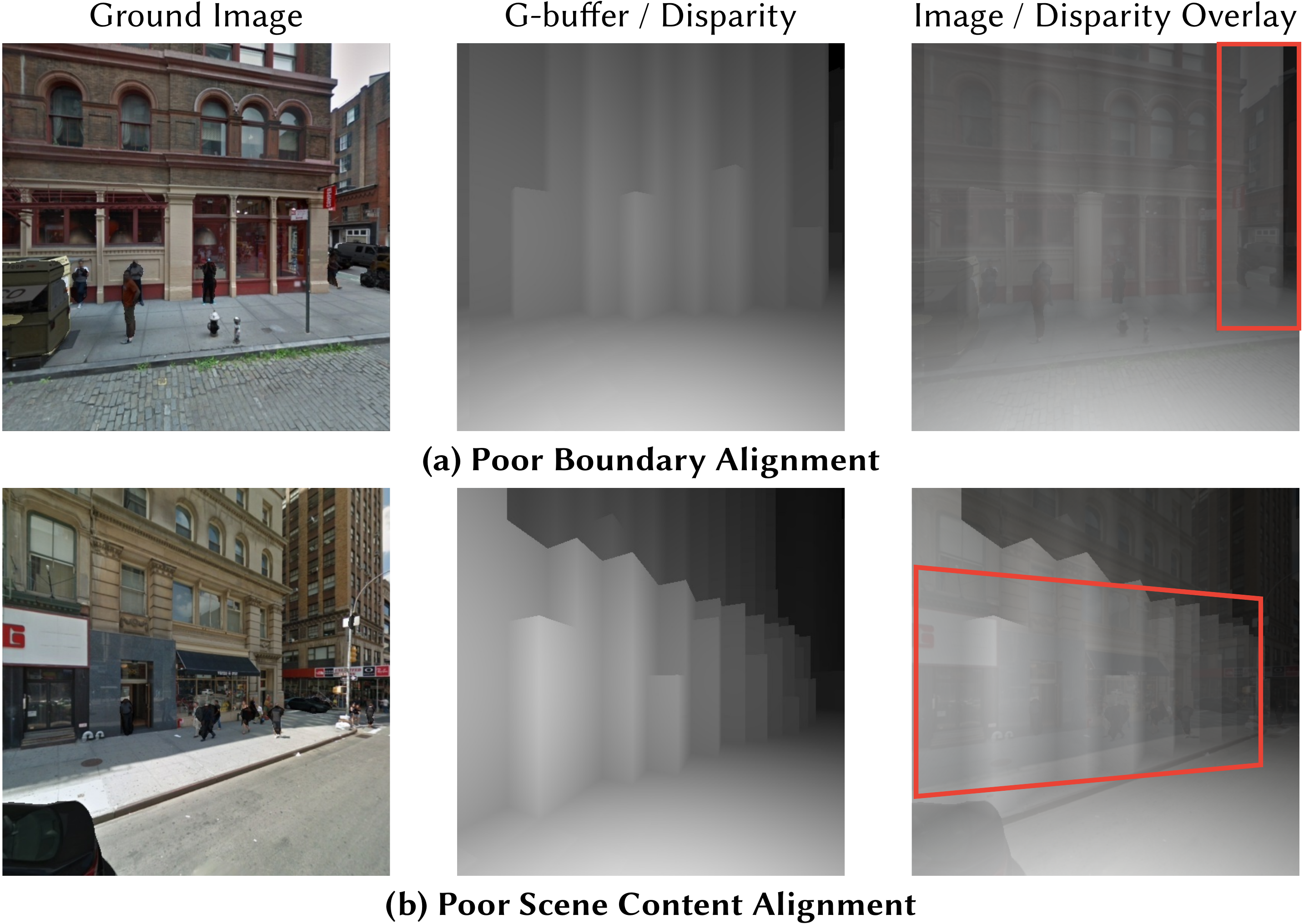}
\caption{
\textbf{Misalignment of Aerial Data and Ground Images}.
We show two representative examples of misalignment of aerial data (height maps) and the ground images in our dataset.
By overlaying disparity maps rendered from the height maps atop the corresponding ground-level RGB views, we observe noticeable misalignment.
Such misalignment includes the offsets near occlusion boundaries that may be attributable to systematic error due to the precision of aerial captures. 
We also see misalignment of scene content, due to the discrepancy between the the static height maps we use and the ever-changing street scenes characteristic of vibrant cities like NYC.
These discrepancies and misalignments necessitate that our system be robust to data noise.
}
\label{fig:data_error}
\end{figure}

%% file: tbls/ablation.tex
\begin{table}
\caption{
\textbf{Ablation on Autoregressive Video Diffusion}
(\colorbox{tbl_no1}{best}, \colorbox{tbl_no2}{$2_\text{nd}$ best}).
We first validate our choices of using imputation against an alternative method where we add extra RGB channels to ControlNet by feeding the condition image to the first frame and a black image to the second frame (\emph{RGB Control}). We find that RGB Control still demonstrates severe quality degeneration after only $16$ steps.
Moreover, we validate the impact of the resampling in temporal imputation (Ours vs Ours - Resample) and the initialisation from a noised image warped from the condition image to the target view (Ours vs Ours - WarpInit).
We notice that both techniques help improve the consistency (lower LPIPS) of generation.
}
\label{tbl:ablation}
\begin{center}
\resizebox{\linewidth}{!}{
\begin{tabular}{lcccc} 
\toprule
\multirow{2}{*}{Metric} &  \multirow{2}{*}{LPIPS $\downarrow$} & \multicolumn{3}{c}{FID$\downarrow$/KID$\downarrow$ @ Steps}\\
 & & 1--16 & 16--32 & 32--64\\
\midrule
RGB Control & $0.603$ & $25.29$/$0.024$ & $96.29$/$0.112$ & $136.68$/$0.153$\\
\midrule
Ours - Resample & $0.525$ & $14.29$/$0.013$ & \cellcolor{tbl_no1}{$25.56$/$0.028$} & \cellcolor{tbl_no1}{$31.02$/$0.030$} \\
Ours - WarpInit & \cellcolor{tbl_no2}{$0.522$} & \cellcolor{tbl_no1}{$11.51$/$0.012$} & $26.57$/$0.029$ & $36.00$/$0.035$ \\
Ours & \cellcolor{tbl_no1}{$0.519$} & \cellcolor{tbl_no2}{$11.69$/$0.012$} & \cellcolor{tbl_no2}{$25.59$/$0.028$} & \cellcolor{tbl_no2}{$35.47$/$0.034$} \\
\bottomrule

\end{tabular}
} 
\end{center}
\end{table}

%% file: figs/ablation_rgbcontrol.tex
\begin{figure*}[ht]
\centering
\includegraphics[width=0.96\linewidth]{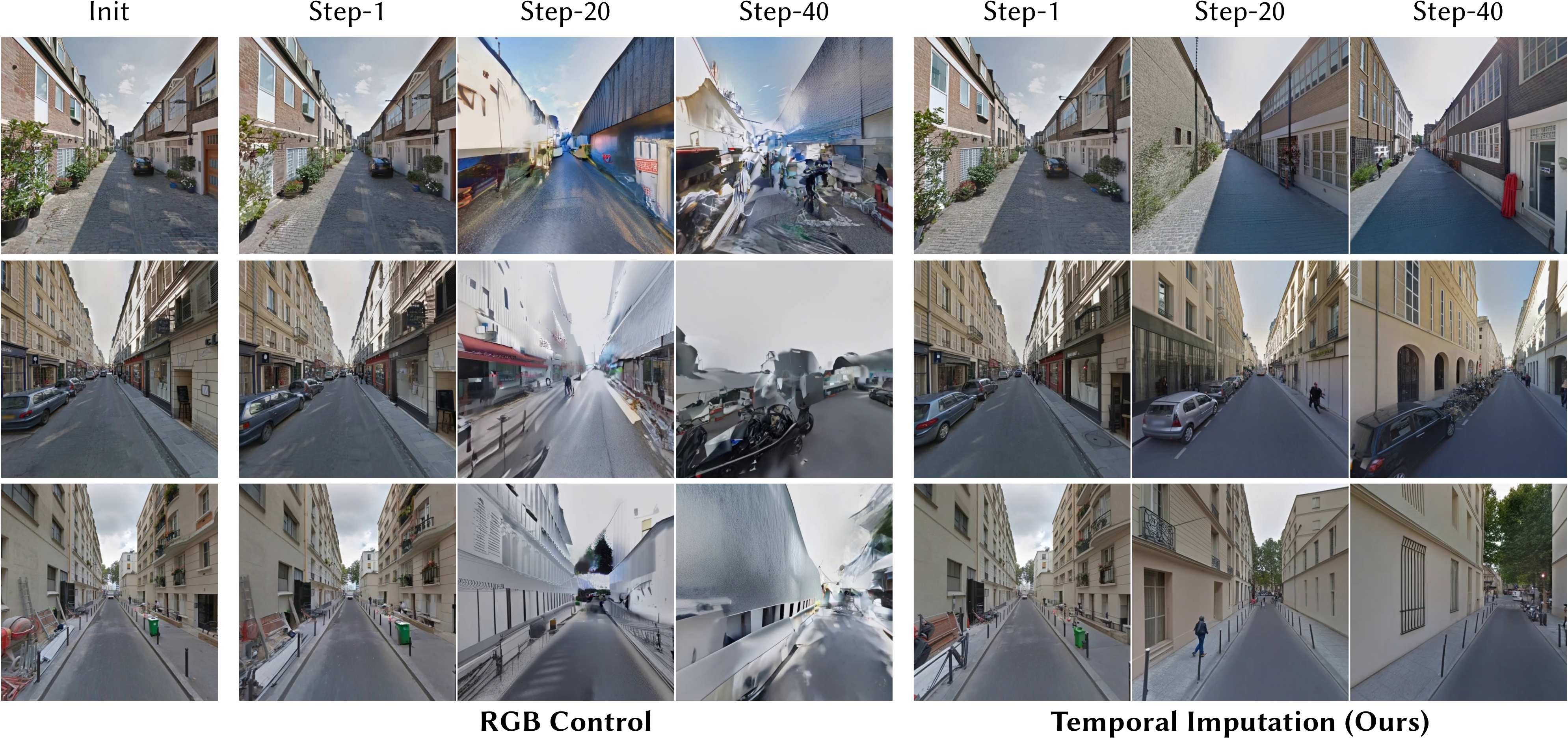}
\caption{
\textbf{Temporal Imputation (Ours) vs RGB Control}.
We compare temporal imputation, our autoregressive video diffusion method, to an alternative method, \emph{RGB Control}, that adds the condition image as extra RGB channels to the ControlNet~\cite{zhang2023adding}.
Both methods are using the same two-frame diffusion model.
We observe that while RGB Control still suffers from serious quality degeneration, our temporal imputation method robustly generates high quality \artifacts across a long range.
}
\label{fig:ablation_rgbcontrol}
\end{figure*}

%% file: figs/ablation_resample_warpinit.tex
\begin{figure*}[ht]
\centering
\includegraphics[width=0.96\linewidth]{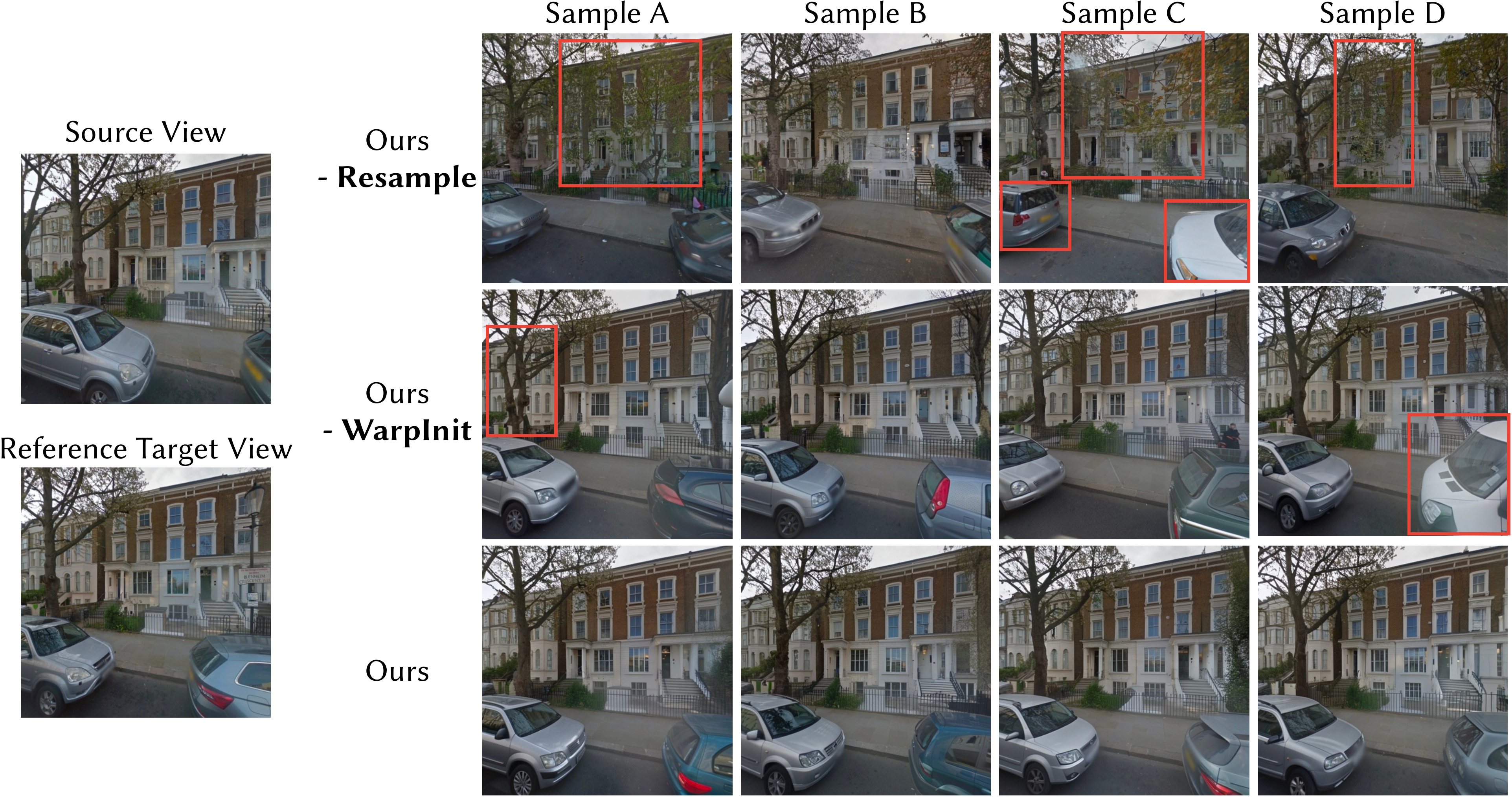}
\caption{
\textbf{Impact of Resample and WarpInit}.
In this experiment, we validate the impact of Resample and WarpInit techniques (see Sec. 3.3) in temporal imputation.
Given the source view on the left, we use different variants of temporal imputation, one without Resample (Ours - Resample), one without WarpInit (Ours - WarpInit), and the full temporal imputation, to autoregressively generate a target view, for a few samples using different random seeds.
Compared to the reference real target view image, we find that our full temporal imputation produces most similar samples, indicating the best consistency in near-range view generation.
}
\label{fig:ablation_resample_warpinit}
\end{figure*}